\documentclass{article}

\usepackage{macro}

\usepackage{arxiv}

\usepackage[utf8]{inputenc} 
\usepackage[T1]{fontenc}    
\usepackage{hyperref}       
\usepackage{url}            
\usepackage{booktabs}       
\usepackage{amsfonts}       
\usepackage{nicefrac}       
\usepackage{microtype}      
\usepackage{lipsum}
\usepackage{graphicx,verbatim,float,subfigure,graphicx,color,siunitx}
\usepackage{amsmath}
\graphicspath{ {./images/} }

\usepackage[utf8]{inputenc} 
\usepackage[T1]{fontenc}    
\usepackage{hyperref}       
\usepackage{url}            
\usepackage{booktabs}       
\usepackage{amsfonts}       
\usepackage{nicefrac}       
\usepackage{microtype}      
\usepackage{xcolor}         
\usepackage[shortlabels]{enumitem}
\usepackage[parfill]{parskip}
\usepackage{amsmath}
\usepackage{dsfont}

\newtheorem{theorem}{Theorem}
\newtheorem{theoremproof}{Proof of Theorem}

\newtheorem{lemma}{Lemma}
\newtheorem{lemmaproof}{Proof of Lemma}

\title{Neural Network approximation power on homogeneous and heterogeneous reaction-diffusion equations}

\author{
  Haotian Feng$^*$\\
  Dept. of Radiation Oncology \\
  University of California-San Francisco \\
  San Francisco, CA 94115  \vspace{0.05in} \\
}

\begin{document}
\vspace{-0.02in}
\maketitle

\begin{abstract}

Reaction–diffusion systems represent one of the most fundamental formulations used to describe a wide range of physical, chemical, and biological processes. With the increasing adoption of neural networks, recent research has focused on solving differential equations using machine learning techniques. However, the theoretical foundation explaining why neural networks can effectively approximate such solutions remains insufficiently explored.

This paper provides a theoretical analysis of the approximation power of neural networks for one- and two-dimensional reaction–diffusion equations in both homogeneous and heterogeneous media. Building upon the universal approximation theorem, we demonstrate that a two-layer neural network can approximate the one-dimensional reaction–diffusion equation, while a three-layer neural network can approximate its two-dimensional counterpart. The theoretical framework presented here can be further extended to elliptic and parabolic equations.

Overall, this work highlights the expressive power of neural networks in approximating solutions to reaction–diffusion equations and related PDEs, providing a theoretical foundation for neural network–based differential equation solvers.

\end{abstract}


\section{Introduction}\label{sec:intro}
The reaction-diffusion equation describes the process where substances or quantities spread through space (diffusion) and local interactions (reaction). It is used to model a wide range of dynamical processes including chemistry, biology and physics, capturing behaviors like population dynamics, chemical reactions, and heat transfer. One prominent example is the Fisher-KPP equation, which models the spread of biological populations through logistic growth and diffusion. Fisher-KPP equation is widely used in modeling the tumor growth and disease spreads, such as the glioblastoma. Compared to medical imaging based analysis, Fisher-KPP equation provides more insights of the physical environment and tumor dynamics, leading to a better prediction of tumor evolution process and a more personalized treatment for patients.

In recent years, machine learning methods especially neural networks have been extended to solving partial differential equation (PDE) related physics, chemistry and engineering problems. Compared to numerical analysis, neural networks excel in handling complex and high-dimensional problems, incorporating flexibility in modeling nonlinearities and reducing computational cost once it is fully trained. One of the known machine learning-based PDE solvers is the Physics-Informed Neural Network (PINN) [1], which utilizes an artificial neural network and embeds physics laws inside the loss function. However, such method is strictly limited on whether we can find an explicit expression into loss function. One of the more fundamental problem is, how can we use Deep Learning to solve complex PDEs and quantitatively estimate the error bound of Neural Network. PDEs are among the most ubiquitous tools used to form fundamental problems in Engineering including Solid Mechanics, Fluid Mechanics and man-made complex control systems. However, many PDEs especially the nonlinear PDEs are pretty hard to solve and we even cannot find the analytical solution, for example, the Navier-Stokes equation. 

The reaction-diffusion equation is a type of hyperbolic partial differential equation that models the behavior of systems in which substances undergo both diffusion and chemical reactions. Reaction-diffusion equation is widely used in modeling biology, chemistry, physics and ecology phenomena. Depending on the expression of reaction term, there are multiple forms of reaction-diffusion equation, including Fisher-KPP Equation used in poulation dynamics like tumor cell modeling, Lotka-Volterra Equation in predator-prey modeling, and Turing Patterns used in morphogenesis modeling. 

\section{Related works}\label{sec:literature}
Machine Learning-based approximation methods for PDEs was first targeted on low-dimensional PDEs in 1994 by Dissanayake[2], where the authors considered several differential operators operated on transfer function like linear 2D Poisson equation and nonlinear 2D thermal conduction equation. The author proved that Neural Network can be used to predict values at certain time step. Later in 1998, Lagaris[3] extended the problem into more general ODEs and PDEs defined in 3D orthogonal box boundaries. The specific solution of the differential equation is calculated based on initial/boundary conditions, while the general solution involves a feed-forward neural network with adjustable parameters (the weights). The authors proved that the solution obtained has an excellent closed analytical form and can be applied to both ODE and PDE with careful choice of trail solution. In 2003 Li[4] proposed a Neural Network by considering the activation function of hidden neurons to be radial basis functions whose parameters are learned by a two-stage gradient descent strategy. Much later in 2017, with the emerging of more advanced deep learning frameworks, E[5] extended DNN to high-dimensional PDEs. The authors formulate PDEs as equivalent backward stochastic differential equations (BSDEs). Then the BSDE is viewed as stochastic control problem analogous to policy function, where such policy function can be approximated with a DNN, just as in reinforcement learning. In this way, the PDE is not formulated with initial value problems, but instead terminal conditions to connect with BSDE. Han[6] further proposed that for high-dimensional PDE, we can reformulate the PDE into numerical form of BSDE and utilize this methods to different kinds of PDEs. Later in recent years, several researches have been conducted to understand how Neural Network can be used to solve different real life problems including in complex geometries[7] and the Kirchhoff plate bending problem[8]. Several research extends solving high-dimensional partial differential equations with deep neural networks [9] [10] and tries to gain theoretical understanding of the deep neural network to solve PDEs [11] [12] [13].

To further understand how the Neural Network performs for solving PDEs, in 2017, Petersen[14] studied the necessary and sufficient complexity of ReLU neural networks - in terms of depth and number of weights - which is required for approximating classifier functions in an $\mathcal{L}^p$-sense. They authors investigate the approximation properties of neural networks by studying how complex networks need to be in order to approximate certain functions well. Based on Petersen's conclusion, in 2018, Sirignano [15] developed a deep learning algorithm to solve high-dimensional PDEs, which is a mesh-free algorithm and showed that the algorithm could solve a class of high-dimensional free boundary PDEs in up to 200 dimensions. Moreover, the authors prove a theorem regarding the approximation power of neural network for quasilinear PDEs. Grohs[16] gives a theoretical proof that Neural Network can overcome the curse of dimensionality of numerical analysis of solving Black-Scholes PDEs.

Previous research has demonstrated the great potential of Neural Networks and Deep Learning in solving complex problems. However, a rigorous theoretical analysis of their approximation power, particularly in the context of partial differential equations (PDEs), remains limited. This study aims to provide a theoretical estimation of the approximation capability of Neural Networks applied to the Fisher–KPP equation, a representative form of the reaction–diffusion equation. Reaction–diffusion equations are widely used to model various biomedical processes and incorporate characteristics of several fundamental PDE classes, such as elliptic (e.g., Poisson’s equation) and parabolic equations. These properties make the reaction–diffusion equation an appropriate and meaningful target for theoretical analysis.

\section{Problem Setup}

\subsection{Domain discretization}\label{sec:discretize}
\begin{figure}[h!]
\vspace{-0.1in}
\centering
\subfigure[]{
  \includegraphics[width=0.4\textwidth]{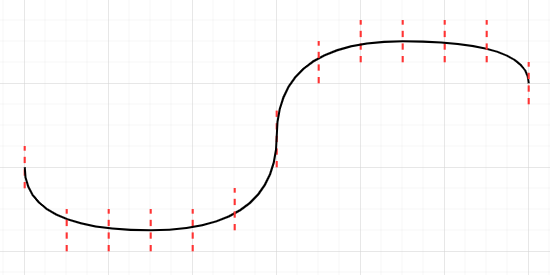}
}
\centering
\subfigure[]{
  \includegraphics[width=0.4\textwidth]{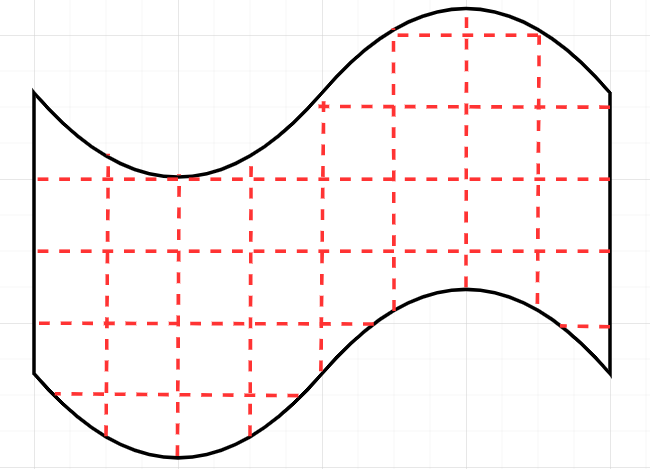}
}
\caption{Domain discretization of (a) 1D arbitrary curve (b) 2D arbitrary domain}
\label{img:domain_discrete}
\end{figure}

The key idea to approximate the PDE solution function with Neural Network is that, for each infinitesimal interval or domain, the analytical expression of curves or surfaces can be described numerically and at the same time, can be approximated by one or a combination of activation functions (like ReLU or Step activation function). An example of domain discretization is shown in Figure~\ref{img:domain_discrete}. For 1D curve, we can split the curve into several small curve segments, while for 2D domain, we can split the domain into several subdomains. Within each discretized subdomain, the first derivative can be numerically represented as: $f'(x)=\frac{f(x+h)-f(x)}{h}=\frac{f(x)-f(x-h)}{h}=\frac{f(x+h)-f(x-h)}{2h}$. Moreover, the second derivative can be represented using central difference formula: $f''(x)=\frac{f(x+h)-2f(x)+f(x-h)}{h^2}$. With this setting, since the whole domain is connected by different subdomains, we will only need to prove that, within each discretized subdomain, the solution of the targeting PDE can be represented by the neural network. Then the whole domain will also be approximated by neural network.

\subsection{Neural Network Approximation Power Theorems}
To prove the Neural Network can be used to approximate solution equation of Partial Differential Equations, we will rely on the Universal Approximation Theorem (UTS) to show the approximation power of Neural Network on functions in 1D and in higher dimensions.

\paragraph{1D Universal Approximation Theorem}
Suppose $g:[0,1)\to \mathcal{R}$ is $\rho$-Lipshitz. For any $\epsilon > 0$, there exists a 2-layer neural network $h(x)$ with 2m neurons where $m:=\lceil\frac{\rho}{\epsilon}\rceil$, such that $\forall x \in [0,1), |f(x)-g(x)|\leq\epsilon$.
\label{theo:1d_nn}

\paragraph{High Dimensional Universal Approximation Theorem}
Suppose $g$ is a continuous function and $\epsilon > 0$. Additionally, assume that $\forall \delta>0, \forall \mathbf{x_1,x_2}\in[0,1]^d$, if $||\mathbf{x_1}-\mathbf{x_2}||_{\infty}\leq\delta$, we have $|g(\mathbf{x_1})-g(\mathbf{x_2})|\leq\epsilon$, Then, there exists a 3-layer neural network $h(\mathbf{x})$ with $\Omega(\frac{1}{\epsilon^d})$ ReLU neurons such that $\forall \mathbf{x}\in [0,1]^d, |h(\mathbf{x})-g(\mathbf{x})|\leq 2\epsilon$.
\label{theo:2d_nn}

The universal approximation theorem can be extended to the discretized domains, as formalized in following lemma.

\begin{lemma}\label{lemma:discrete}
    For a continuous solution $u=f(x)$ in the domain of interest: when the domain is discretized into multiple connected subdomains, and the solution in each subdomain satisfies UTS, then the solution in whole domain also satistfies the UTS.
\end{lemma}

\begin{lemmaproof}
(1) 1D Case: For the 1D case, we only to prove Lipshitz continuous condition can be extended from discrete domain to whole domain. Assuming a 1D domain $\Omega$ discretized into N different subdomains. Each subdomain $i$ is denoted as $N_i$ and is $\rho_i$-Lipshitz by definition. Two adjacent subdomains $\Omega_i$ and $\Omega_j$ are smoothly connected through the edge $\partial \Omega_{ij}$. Considering any two points \textbf{$x_1 \in \Omega_1$} and \textbf{$x_2 \in \Omega_2$}, and a point \textbf{$x_{12} \in \partial \Omega_{12}$} which is the intersection point between edge and straight line formed by $x_1$ and $x_2$. By the definition of Lipshitz, we have:

\vspace{-0.01in}
\begin{equation}
\begin{aligned}
|f(x_2)-f(x_1)| &= |f(x_2) - f(x_{12}) + f(x_{12}) - f(x_1)| \\
&\leq |f(x_2) - f(x_{12})| + |f(x_{12}) - f(x_1)| \\
&\leq \rho_2||x_2-x_{12}|| + \rho_1||x_{12}-x_1|| \\
&\leq max(\rho_1, \rho_2)||x_2-x_1||
\end{aligned}   
\label{eqn:laplace_1}
\end{equation}

That is, the $\rho-$ Lipshitz continuity is extended to two connected subdomains. Thus, we could extend the conclusion to the whole domain, where the whole domain is $max(\rho_i)$ Lipshitz continuous.

(2) High-Dimensional Case: For the high-dimensional space, each variable $x$ is represented as a d-dimensional vector $R^d$. Similarly consider two points in two adjacent subdomains in the d-dimensional hyper-rectangle space \textbf{$x_1 \in \Omega^d_1$} and \textbf{$x_2 \in \Omega^d_2$}, and the intersection point \textbf{$x_{12} \in \partial \Omega^d_{12}$}. From the definition, we know the infinity norm between $x_1$, $x_2$ and $x_{12}$ are bounded.

\vspace{-0.01in}
\begin{equation}
\begin{aligned}
|f(x_2)-f(x_1)| &= |f(x_2) - f(x_{12}) + f(x_{12}) - f(x_1)| \\
&\leq |f(x_2) - f(x_{12})| + |f(x_{12}) - f(x_1)| \\
&\leq \epsilon_1 + \epsilon_2 \\
&= \epsilon
\end{aligned}   
\label{eqn:laplace_2}
\end{equation}

Similarly, the conclusion can be extended to the whole domain. Lemma is proved.

\end{lemmaproof}

\subsection{Reaction-Diffusion Equation}\label{sec:rd_eqn}
In this paper, Fisher-KPP is consdered as the target reaction-diffusion equation. The general expression of Fisher-KPP equation is $\frac{\partial u}{\partial t} = D\nabla^2u+ru(1-u)$, where u represents the population density, D is the diffusion coefficient and r is the growth rate of the population. $ru(1-u)$ is the reaction term, which is limited by the carrying capacity to normalize the u between 0 and 1. When u is much less than 1, the reaction term behaves like $ru$ and u grows exponentially. When u is close to 1, the population stabilizes at the carrying capacity. 

\paragraph{Homogeneous Setting}
The homogeneous reaction-diffusion equation follows the expression as below, where x is the independent parameter vector. 
\vspace{-0.01in}
\begin{equation}
\begin{split}
&\frac{\partial u}{\partial t} = D\Delta u + ru(1-u) \quad in \quad \Omega \\
&u(\Bar{\textbf{x}}) = g(\Bar{\textbf{x}}) \quad on \quad \partial \Omega_g \\
&u(\Bar{\textbf{x}}) = h(\Bar{\textbf{x}}) \quad on \quad \partial \Omega_h \\
\end{split}   
\label{eqn:laplace}
\end{equation}

where $\Delta$ is the Laplace operator and $\partial \Omega_g$ is the Dirichlet boundary condition and $\partial \Omega_h$ is the Neumann boundary condition. $\Bar{\textbf{x}}$ refers to the target point in physical space, which is a vector of d-dimension.

\paragraph{Heterogeneous Setting}
The heterogeneous reaction-diffusion equation follows the following expression, similarly, where x is the vector of independent parameters. 
\vspace{-0.01in}
\begin{equation}
\begin{split}
& \frac{\partial u}{\partial t} = \nabla\cdot (D(\Bar{\textbf{x}})\nabla u) + ru(1-u) \quad in \quad \Omega \\
&u(\Bar{\textbf{x}}) = g(\Bar{\textbf{x}}) \quad on \quad \partial \Omega_g \\
&u(\Bar{\textbf{x}}) = h(\Bar{\textbf{x}}) \quad on \quad \partial \Omega_h \\
\end{split}
\label{eqn:nonhomo}
\end{equation}

where $D(\Bar{\textbf{x}})$ models different values at different spatial locations and introduces the heterogeneity to the equation.

\subsection{Assumptions}
To formalize the proving process, there are several assumptions need to be added:
\begin{itemize}
    \item The domain of interest is finitely large, that is, there exists a boundary that fully encloses the domain.
    \item There exists a smooth solution for reaction-diffusion equation. 
    \item The boundary conditions are bounded. That is, the dirichlet boundary condition $g(x)$ and the Neumann boundary condition $h(x)$ are bounded. 
\end{itemize}

\section{Theoretical Analysis of Neural Network's approximation power}
Since the time-dependent term in the Fisher-KPP equation can be approximated with a series of networks, the prediction power can be validated considering the time-independent case, which reduces the hyperbolic reaction-diffusion equation to the elliptic reaction-diffusion equation. Before the main theorem, we first propose several lemmas to aid the proof the main theorem.

Given the fact that the value of u is bounded between 0 and 1 by problem setting, the reaction term is also bounded. Thus we have the following Lemma:

\begin{lemma}\label{lemma:lip}
    The first-order derivative of Fisher-KPP's solution is Lipshitz continuous. 
\end{lemma}

\begin{lemmaproof}
    
To prove the lemma, first consider the homogeneous expression of Fisher-KPP: $D\Delta u + ru(1-u) = 0$ in d-dimensional space, that is, $u, x \in R^d$. Considering a hyper-rectangle subdomain $N_i$ in the domain of interests, for any choices of $x_1 < x_2$, we have $\frac{u'(x_2)-u'(x_1)}{x_2-x_1} + ru(1-u)=0$. where $u\in[0,1]$, the reaction term $ru(1-u)\in[0,\frac{r}{4}]$. Thus, we know $\frac{u'(x_2)-u'(x_1)}{x_2-x_1} \in [-\frac{r}{4}, 0]$. Consequently, $|u'(x_2)-u'(x_1)|\leq \frac{r}{4}|x_2-x_1|$. Thus, the first-order derivative of Fisher-KPP's solution u is Lipschitz-continuous.

Next move to the heterogeneous expression of Fisher-KPP: $\nabla\cdot (D(\Bar{\textbf{x}})\nabla u) + ru(1-u) = 0$. Similarly, consider a subdomain $N_j$ and acknowledging that reaction term is bounded. For any point $x$ in the subdomain, we could choose two adjacent points $x_1=x-h < x_2=x+h$ where h is very small. From the setting, we know that in any arbitrary dimension i, when $h_i\to 0$, the known diffusion coefficient function $D(x)$ has $D(x+h)\approx D(x-h)$. Thus we could rewrite the heterogeneous diffusion term as:

\begin{equation}
\begin{aligned}
\nabla \cdot (D(\bar{\mathbf{x}})\nabla u) 
&= \nabla D \cdot \nabla u + D\nabla^2 u \\
&= \sum_{i=1}^{d} \Bigg[
    \frac{D(x_i+h_i)-D(x_i-h_i)}{2h_i}
    \cdot
    \frac{u'(x_i+h_i)+u'(x_i-h_i)}{2} \\
&\quad + \frac{D(x_i+h_i)+D(x_i-h_i)}{2}
    \cdot
    \frac{u'(x_i+h_i)-u'(x_i-h_i)}{2h_i}
\Bigg] \\
&= \sum_{i=1}^{d} \frac{1}{2h_i}
   \big[D(x_i+h_i)u'(x_i+h_i) - D(x_i-h_i)u'(x_i-h_i)\big] \\
&= \sum_{i=1}^{d} \frac{D(x_i)}{2h_i}
   \big[u'(x_i+h_i)-u'(x_i-h_i)\big]
\end{aligned}
\end{equation}

By equation definition, we have $\sum_{i=1}^{d} \frac{D(x_i)}{2h_i}(u_i'(x_i+h_i)-u_i'(x_i-h_i)) = -ru(1-u) \in [-\frac{r}{4}, 0]$. reshaping the function, we have $|u'(x+h)-u'(x-h)|=-ru(1-u)\frac{2h}{D(x)} <= |\frac{(x+h)-(x-h)}{D(x)}|\cdot|-ru(1-u)| <= \frac{r}{4D_{min}}|(x+h)-(x-h)|$. That is, $|u'(x_2)-u'(x_1)| <= \frac{r}{4D_{min}}|x_2-x_1|$, the lemma is proved.
\end{lemmaproof} 

From the conclusion of Lemma~\ref{lemma:lip}, we could easily conclude the following:
\begin{lemma}\label{lemma:bnd}
    The first-order derivative of Fisher-KPP's solution is bounded.
\end{lemma}

\begin{lemmaproof}
    Assuming the first-order derivative of Fisher-KPP's solution is $\phi'(x)$, at any point of interest $x\in \Omega$. Based on Assumption 2, we know that for any $x_h\in \partial\Omega$, $\phi'(x_h)$ is bounded. For any point in the domain $\Omega$, from Lemma 2, we know $|\phi'(x)-\phi'(x_h)|\leq C_1||x-x_h||$.

    From Assumption 1, we know $||x-x_h||$ is bounded by a constant $C_2$ for an enclosed domain, thus, $|\phi'(x)-\phi'(x_h)|\leq C_1*C_2=C$. As $\phi'(x_h)$ is bounded, so for any $x \in \Omega$, $\phi'(x)$ must be bounded. Lemma proved.
\end{lemmaproof}

\subsection{Approximation of Fisher-KPP Equation}

After confirming the Lipshitz continuity in previous lemma, we could continue to apply the universal approximation theorem to the Fisher-KPP equation. This section discusses how neural networks can be used to approximate 1D and 2D equations. The proof is structured with the following steps: (1) Prove that at each time step T, u(x,t=T) is Lipschitz continuous. (2) Prove that the time-dependent term is Lipschitz-continuous. (3) Prove the whole term u(x,t) is Lipshitz continuous. (4) Apply universal approximation theorem under Lipshitz continuous condition.



\subsubsection{Approximation on 1D Fisher-KPP Equation}
This section illustrates that the solution of 1D Fisher-KPP equation can be approximated by a 2-layer Neural Network utilizing the previous theorems and lemmas. 

\begin{theorem}
Suppose function $g$ is the solution from 1D Fisher-KPP Equation. For any $\epsilon>0$, there exists a 2-layer neural network $h(x)$ with $2m$ neurons where $m=\frac{\rho'}{\epsilon}$, such that $\forall x \in [0,1)$, $|h(x)-g(x)|\leq \epsilon$. Specifically, bounded term $|x|\leq\delta$ and $\rho'=C(c_0+2\delta+h)$, 
\label{theo:1d_lap}
\end{theorem}

\begin{theoremproof}

In order to prove the theorem, as mentioned above, we want to first prove that the solution function $g$ is a $\rho-$ Lipschitz function and then prove that it can be approximated by a 2-layer neural network.

First, the 1D domain of interest is divided into \textbf{m} different subdomains, where each domain $j$ is bounded by the range $[x_j,x_{j+1}]$, where $x_j=\frac{j-1}{m}$ and $\|x_{j+1}-x_j\| \in [0,1)$. Within individual subdomain, the numerical representations in Section~\ref{sec:discretize} will hold. That is, for any points $x_1<h<x_2$ in subdomain, where $h$ lies between $x1$ and $x2$. We have $g(h)=g(x_2)-g'(x_2)(x_2-h)=g(x_1)+g'(x_1)(h-x_1)$. Moreover, from Lemma~\ref{lemma:lip} and~\ref{lemma:bnd}, we know that $g'(x)$ is $\rho-$Lipshitz continuous and bounded by constant C. Consequently, 

\begin{equation}
\begin{aligned}
|g(x_2)-g(x_1)| &= |g'(x_2)(x_2-h) + g'(x_1)(h-x_1)| \\
&= |x_2g'(x_2)-x_1g'(x_1)-h(g'(x_2)-g'(x_1))| \\
&\leq |x_2g'(x_2)-x_1g'(x_1)| + |h(g'(x_2)-g'(x_1))| \\
&\leq |x_2g'(x_2)-x_2g'(x_1)| + |x_2g'(x_1)-x_1g'(x_1)| + |h||(g'(x_2)-g'(x_1))| \\
&\leq \rho|x_2||x_2-x_1| + C|x_2-x_1| + \rho h|x_2-x_1| \\
&\leq (C+\rho+h\rho)|x_2-x_1| \\
&= \rho'|x_2-x_1|
\end{aligned}
\end{equation}

Thus function g is $\rho'-$Lipshitz. Using 1D Universal Approximation Theorem, we know that there exists a 2-layer neural network with $2\lceil\frac{\rho'}{\epsilon}\rceil$ neurons, such that the error bound $|h(x)-g(x)|\leq\epsilon$.

\end{theoremproof}

\subsubsection{Approximation on high dimensional Fisher-KPP Equation}\label{sec:2d_poisson}

\begin{theorem}
Suppose function g is the solution for high-dimensional Fisher-KPP equation. Then there exists a 3-layer neural network h(x) with ReLU activation function such that $\forall \delta>0, \forall x_1,x_2$ satisfying if $||x_1-x_2||_{\infty}\leq\frac{\delta}{m}$, we have $|h(x)-g(x)|\leq\epsilon$. 
\label{theo:2d_lap}
\end{theorem}

\begin{theoremproof}

First, the high-dimensional domain is discretized of \textbf{m} subdomains (each domain can be viewed as small hyper-rectangles), where the $i\times j$-th domain can be represented as $[x_i,x_{i+1}]\times[y_j,y_{j+1}]$. Here $x_i=\delta \frac{i}{m}$ and $y_j=\delta \frac{j}{m}$. Then the second order derivative of any function $\phi(x,y)$ can be numerically computed using central difference formula, as shown in Equation~\ref{eqn:central_difference}.
\begin{equation}
    \Delta\phi(x,y)=\frac{\phi_{i-1,j}-2\phi_{i,j}+\phi_{i+1,j}}{h^2} + \frac{\phi_{i,j-1}-2\phi_{i,j}+\phi_{i,j+1}}{h^2}
\label{eqn:central_difference}
 \end{equation}

Then, from $|\phi'|\leq c_\delta$ we can derive that $\frac{\phi(x+h)-\phi(x)}{h}\leq c_\delta$, for $h\to 0$. Then we can rewrite the Fisher-KPP Equation's diffusion term as Equation~\ref{eqn:LE_rewrite}.
\begin{equation}
\begin{aligned}
    |h^2f(i,j)|&=|\phi_{i-1,j}+\phi_{i+1,j}+\phi_{i,j-1}+\phi_{i,j+1}-4\phi_{i,j}|\\
    &\leq |\phi_{i-1,j}-\phi_{i,j}|+|\phi_{i+1,j}-\phi_{i,j}|+|\phi_{i,j-1}-\phi_{i,j}|+|\phi_{i,j+1}+\phi_{i,j}|\\
    &\leq 4hc_\delta
\end{aligned}
\label{eqn:LE_rewrite}
\end{equation}

Then for $\forall \Bar{x_1},\Bar{x_2}$, we want to bound the difference between $g(x_1,y_1)$ and $g(x_2,y_2)$. Comparing to Equation~\ref{eqn:LE_rewrite}, $x_1$ is equivalent to position $x_{i-1,j-1}$ and $x_2$ is equivalent to position $x_{i+1,j+1}$. The middle point $x_m$ between $x_1$ and $x_2$ is equivalent to position $x_{i,j}$ and $x_2-x_m=x_m-x_1=h$. Consequently we have:
\begin{equation}
\begin{aligned}
&\quad \,\,2|g(x_2,y_2)-g(x_1,y_1)|\\
&=|2g(x_2,y_2)-g(x_2,y_1)+g(x_2,y_1)-g(x_1,y_2)+g(x_1,y_2)-2g(x_1,y_1)|\\
&\leq |g(x_2,y_2)-g(x_2,y_1)|+|g(x_2,y_1)-g(x_1,y_1)|+|g(x_2,y_2)-g(x_1,y_2)|+|g(x_1,y_2)-g(x_1,y_1)| \\
&\leq |g(x_2,y_2)-g(x_2,y_m)|+|g(x_2,y_m)-g(x_2,y_1)|+|g(x_2,y_1)-g(x_m,y_1)|+|g(x_m,y_1)-g(x_1,y_1)|\\
&+|g(x_2,y_2)-g(x_m,y_2)|+|g(x_m,y_2)-g(x_1,y_2)|+|g(x_1,y_2)-g(x_1,y_m)|+|g(x_1,y_m)-g(x_1,y_1)|\\
&\leq 8hc_\delta:=\epsilon
\end{aligned}
\end{equation}

Thus, we know that for $||x_1-x_2||\leq\frac{\delta}{m}$ and $||y_1-y_2||\leq\frac{\delta}{m}$, we have $||g(\Bar{x_2})-g(\Bar{x_1})||=|g(x_2,y_2)-g(x_1,y_1)|\leq\epsilon$. Here, note that since the solution for Poisson's Equation is smooth, so the maximum slope of solution equation $c_\delta$ is not large. Also the two points here are set to be close in the Theorem 3, so the value of $\epsilon$ is actually small. Finally, we can utilize Theorem~\ref{theo:2d_nn} to prove that there indeed exists a 3-layer Neural Network to approximate the 2D Poisson's Equation.


%

\end{theoremproof}

Thus the general solution $\phi_g(x,y)$ of 2D Poisson's Equation can be approximated by a 3-layer Neural Network. One thing to note that, the constraint $g'(x,y)$ is bounded is actually not a hard constraint on the solution function since the solution for Poisson's Equation is in general a smooth function and has its own physical quantity representation, thus the derivative of the solution function also exists and should be bounded.

\textit{Note: In the theorem, the requirement that \textbf{first derivative of the solution equation should be bounded by a constant $c_\delta$} might not be necessary and may be derived based on the PDE setting, but could depend on the problem setting.}

\textit{For example, we can claim that the first derivative of the solution equation should be bounded by a brief reasoning. We know $\frac{\partial^2 \phi}{\partial x^2}+\frac{\partial^2 \phi}{\partial y^2}=f(x,y)$, this will tell us that when we take the integral over x and y respectively, we could have $\frac{\partial \phi)}{\partial x}+x\frac{\partial^2 \phi}{\partial y^2}=\int f(x,y)dx=C(y)$ and $y\frac{\partial^2 \phi)}{\partial x^2}+\frac{\partial \phi}{\partial y}=\int f(x,y)dy=D(x)$. As we know $\frac{\partial^2 \phi}{\partial x^2}$ and $\frac{\partial^2 \phi}{\partial y^2}$ should be bounded to have physical meaning at any arbitrary point, while function $C(y)$ and $D(x)$ should also be bounded for a bounded region, so we will know that the first derivatives $\frac{\partial \phi}{\partial x}$ and $\frac{\partial \phi}{\partial x}$ should be bounded. \textbf{Thus the requirement that solution equation should be bounded is actually not necessary in structural analysis problems, but might be necessary for other research areas.} }.

\section{Conclusions}
This study focuses on a fundamental class of partial differential equations (PDEs)—the reaction–diffusion equation—which can be readily extended to other types of PDEs. By discretizing the target domain into smaller subdomains where both the numerical representation of derivatives and the activation function approximations hold, the results demonstrate that a two-layer neural network can effectively approximate the solutions of 1D reaction-diffusion Equation. Furthermore, a three-layer neural network shows good approximation performance when the PDE is extended to two-dimensional domains.

Future work can be extended in several directions:
\begin{enumerate}
\item Validate the approximation capability of neural networks for 2D non-homogeneous elliptic equations.
\item Analyze the approximation power of neural networks for reaction–diffusion equations in higher dimensions, such as in 3D space.
\item Investigate whether a more compact neural network architecture—with reduced width or depth—can still provide accurate approximations of PDE solutions.
\end{enumerate}

\section*{References}

{
\small

\begin{enumerate}[ {[}1{]} ]
    
    \item Mao, Zhiping, Ameya D. Jagtap, and George Em Karniadakis. "Physics-informed neural networks for high-speed flows." Computer Methods in Applied Mechanics and Engineering 360 (2020): 112789.
    \item Dissanayake, M. W. M. G., and Nhan Phan‐Thien. "Neural‐network‐based approximations for solving partial differential equations." communications in Numerical Methods in Engineering 10.3 (1994): 195-201. 
    \item Lagaris, Isaac E., Aristidis Likas, and Dimitrios I. Fotiadis. "Artificial neural networks for solving ordinary and partial differential equations." IEEE transactions on neural networks 9.5 (1998): 987-1000.
    \item Jianyu, Li, et al. "Numerical solution of elliptic partial differential equation using radial basis function neural networks." Neural Networks 16.5-6 (2003): 729-734.
    \item Han, Jiequn, and Arnulf Jentzen. "Deep learning-based numerical methods for high-dimensional parabolic partial differential equations and backward stochastic differential equations." Communications in Mathematics and Statistics 5.4 (2017): 349-380.
    \item Han, Jiequn, Arnulf Jentzen, and E. Weinan. "Solving high-dimensional partial differential equations using deep learning." Proceedings of the National Academy of Sciences 115.34 (2018): 8505-8510.
    \item Berg, Jens, and Kaj Nyström. "A unified deep artificial neural network approach to partial differential equations in complex geometries." Neurocomputing 317 (2018): 28-41.
    \item Guo, Hongwei, Xiaoying Zhuang, and Timon Rabczuk. "A deep collocation method for the bending analysis of Kirchhoff plate." arXiv preprint arXiv:2102.02617 (2021).
    \item Raissi, Maziar. "Forward-backward stochastic neural networks: Deep learning of high-dimensional partial differential equations." arXiv preprint arXiv:1804.07010 (2018).
    \item Beck, Christian, et al. "Solving the Kolmogorov PDE by means of deep learning." Journal of Scientific Computing 88.3 (2021): 1-28.
    \item Kutyniok, Gitta, et al. "A theoretical analysis of deep neural networks and parametric PDEs." Constructive Approximation 55.1 (2022): 73-125.
    \item Ruthotto, Lars, and Eldad Haber. "Deep neural networks motivated by partial differential equations." Journal of Mathematical Imaging and Vision 62.3 (2020): 352-364.
    \item Raissi, Maziar, Paris Perdikaris, and George E. Karniadakis. "Physics-informed neural networks: A deep learning framework for solving forward and inverse problems involving nonlinear partial differential equations." Journal of Computational physics 378 (2019): 686-707.
    \item Petersen, Philipp, and Felix Voigtlaender. "Optimal approximation of piecewise smooth functions using deep ReLU neural networks." Neural Networks 108 (2018): 296-330.
    \item Sirignano, Justin, and Konstantinos Spiliopoulos. "DGM: A deep learning algorithm for solving partial differential equations." Journal of computational physics 375 (2018): 1339-1364.
    \item Grohs, Philipp, et al. "A proof that artificial neural networks overcome the curse of dimensionality in the numerical approximation of Black-Scholes partial differential equations." arXiv preprint arXiv:1809.02362 (2018).
\end{enumerate}

}

\newpage
\section*{Appendix}
\addcontentsline{toc}{section}{Appendices}
\renewcommand{\thesubsection}{\Alph{subsection}}
\renewcommand{\thefigure}{B.\arabic{figure}}
\setcounter{figure}{0}

\subsection{Proof of 1D Universal Approximation Theorem}
\begin{theoremproof}
The proof idea is utilizing partition. Target function is almost constant on individual small intervals. So we can construct a function by assigning left hand side value of the target function in the interval. Let $b_i=\frac{i-1}{m}$, where $i=1\cdots m$. Consider
\begin{equation}
    h(x)=\sum_{j=1}^{m}g(b_j)\mathbf{1}_{x\in[b_j,b_{j+1})}
\label{eqn:hx}
\end{equation}

Then we need to replace the indicator function by neurons. Consider step activation function (threshold function),
\begin{equation}
   \sigma(z)=
    \begin{cases}
        0, & z<0 \\
        1, & z\ge 0
    \end{cases}
\end{equation}

Then we can have:
\begin{equation}
    \sigma(z-b_j)-\sigma(z-b_{j+1})=\mathbf{1}_{z\in[b_j,b_{j+1})}
\end{equation}

Thus $h(x)$ in Equation~\ref{eqn:hx} is a 2-layer Neural Network with in total 2m neurons. Then the error bound is given by:
\begin{equation}
    |h(x)-g(x)|=|g(b_j)-g(x)|\le\rho|b_j-x|\leq\rho\cdot\frac{\epsilon}{\rho}=\epsilon
\end{equation}

\end{theoremproof}

\subsection{Proof of 2D Universal Approximation Theorem}
To prove this theorem, we will need to use another lemma that constructs an intermediate piecewise constant function.

\begin{lemma}
Let $g,\delta,\epsilon$ be defined as Theorem~\ref{theo:2d_lap}. For any partition P of $[0,1]^d$ into hyper rectangles, $P={R}_{i=1}^N$ with side length $\leq \delta$, there exists a piece-wise constant function $h(x)=\sum_{i=1}^N \alpha_i \mathbf{1}_{[x\in R_i]}$ such that $\forall x\in [0,1]^d, |h(x)-g(x)|\leq\epsilon$.  
\label{lem:2d_lap}
\end{lemma}

This lemma can be easily approved by assuming $\alpha_i$ be the value of g on any points in the region $R_i$. With this Lemma, we can prove Theorem~\ref{theo:2d_nn}.

\begin{theoremproof}
Divide the domain $[0,1]^d$ into sufficiently small hyper-rectangles of the form $R_i:=\times_{j=1}^d[a_{ij},b_{ij})$. If we can approximate $\mathbf{1}_{[x_j\in[a_{ij},b_{ij})]}$ by 1 layer of the Neural Network with ReLU activation function and approximate $\mathbf{1}_{[\mathbf{x}\in R_i]}$ by 2 layers of the Neural Network with ReLU activation function then we have three layers of the Neural Network as:
\begin{equation}
    f(\mathbf{x})=\sum_i \alpha_i\Tilde{1}_{[\mathbf{x}\in R_i]}
\end{equation}
where $\Tilde{1}$ is approximated by Neural Network with ReLU activation function. Then we have
\begin{equation}
   \mathbf{E}_x|f(x)-g(x)|\leq\mathbf{E}_x|f(x)-h(x)|+\mathbf{E}_x|h(x)-g(x)|\leq 2\epsilon
\label{eqn:expect}
\end{equation}

Now we will construct our two-layer Neural Network to satisfy the above property. First, we will approximate $\mathbf{1}_{x\in[a_{ij},b_{ij}]}$ by the Neurons with ReLU activation function. Consider
\begin{equation}
    \frac{\sigma(z-c_1)-\sigma(z-c_2)}{c_2-c_1}
\label{eqn:relu_2}
\end{equation}

for $c_2>c_1$, as shown in Figure~\ref{img:ReLU}(a). Then Equation~\ref{eqn:relu_2} can form an approximation function of indicator function by
\begin{equation}
    f_{i,j,\gamma}=\frac{1}{\gamma}[\sigma(x_j-a_{ij})-\sigma(x_j-(a_{ij}-\gamma))]-\frac{1}{\gamma}[\sigma(x_j-(b_{ij}+\gamma))-\sigma(x_j-b_{ij})]:=\Tilde{1}_{[x_j\in[a_{ij},b_{ij}]]}
\end{equation}

Note that the $\gamma$ can be chosen to be sufficiently small so that we have as small approximation error of $\Tilde{\mathbf{1}}$ on $\mathbf{1}$ as desired. $f_{i,j,\gamma}(\mathbf{x})$ can only approximate one dimension. To approximate d dimensions, we can compose these 1-layer single-coordinate selector functions to form a 2-layer selector function that selects rectangles in all coordinates:
\begin{equation}
    \Tilde{1}_{[\mathbf{x}\in R_i]}:=f_{i,\gamma}(\mathbf{x})=\sigma(\sum_{j=1}^d f_{i,j,\gamma}(\mathbf{x})-(d-1))
\end{equation}

Then it satisfies that:
\begin{equation}
    f_{i,\gamma}(\mathbf{x})
    \begin{cases}
    1, & \mathbf{x}\in R_i \\
    [0,1)], & o.w. \\
    0, & \mathbf{x}\not\in\times_{j=1}^d
[a_{ij}-\gamma,b_{ij}-\gamma]   
\end{cases}
\end{equation}

Finally, we pick a sufficiently small $\gamma$. By Equation~\ref{eqn:expect}, we conclude that a 3-layer Neural Network with ReLU activation function can approximate high-dimensional Lipshitz function family with small approximation error.

\begin{figure}[h!]
\vspace{-0.1in}
\centering
\subfigure[]{
  \includegraphics[width=0.4\textwidth]{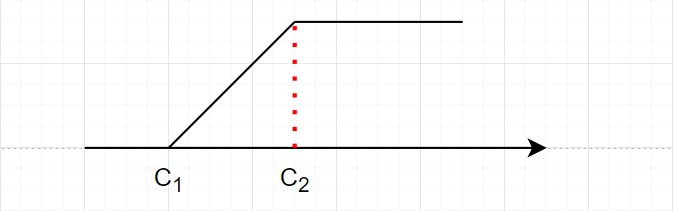}
}
\centering
\subfigure[]{
  \includegraphics[width=0.4\textwidth]{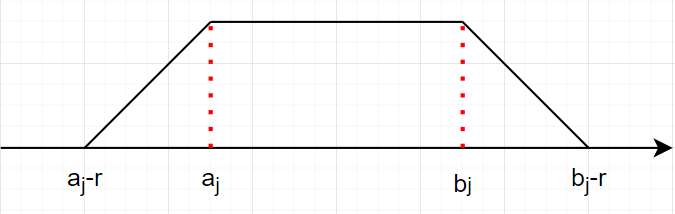}
}
\caption{(a) 1-side ReLU function (b) 2-side ReLU function}
\label{img:ReLU}
\end{figure}

\end{theoremproof}

\end{document}